\documentclass[lettersize,journal]{IEEEtran} 
\usepackage{amsmath,amsfonts}
\usepackage{algorithm}
\usepackage{algpseudocode}
\usepackage{array}
\usepackage[caption=false,font=normalsize,labelfont=sf,textfont=sf]{subfig}
\usepackage{textcomp}
\usepackage{stfloats}
\usepackage{url}
\usepackage{verbatim}
\usepackage{graphicx}
\usepackage{cite}
\usepackage{amsmath}
\usepackage{amssymb}
\usepackage{xcolor}
\usepackage{booktabs}
\usepackage{multirow}
\usepackage{bbding}
\usepackage{orcidlink}
\usepackage{bbm}
\usepackage{colortbl}
\usepackage[capitalise]{cleveref}
\begin{document}

\title{FCOS: A Two-Stage Recoverable Model Pruning Framework for Automatic Modulation Recognition}

\author{Yao Lu\orcidlink{0000-0003-0655-7814}, ~\IEEEmembership{Student Member, IEEE}, Tengfei Ma, Zeyu Wang, Zhuangzhi Chen, Dongwei Xu\orcidlink{0000-0003-2693-922X},~\IEEEmembership{Member, IEEE}, Yun Lin\orcidlink{0000-0003-1379-9301},~\IEEEmembership{Senior Member, IEEE}, Qi Xuan\orcidlink{0000-0002-6320-7012},~\IEEEmembership{Senior Member,~IEEE}, Guan Gui,~\IEEEmembership{Fellow,~IEEE}
\thanks{This work was partially supported by the Key R\&D Program of Zhejiang under Grant 2022C01018 and by the National Natural Science Foundation of China under Grant U21B2001, 62301492 and 61973273. (Corresponding author: Qi Xuan, Dongwei Xu)}
\thanks{Yao Lu is with the Institute of Cyberspace Security, College of Information Engineering, Zhejiang University of Technology, Hangzhou 310023, China, with the Binjiang Institute of Artificial Intelligence, Zhejiang University of Technology, Hangzhou 310056, China, also with the Centre for Frontier AI Research, Agency for Science, Technology and Research, Singapore 138632 (e-mail: yaolu.zjut@gmail.com).}
\thanks{Tengfei Ma, Zeyu Wang, Dongwei Xu and Qi Xuan are with the Institute of Cyberspace Security, College of Information Engineering, Zhejiang University of Technology, Hangzhou 310023, China, also with the Binjiang Institute of Artificial Intelligence, Zhejiang University of Technology, Hangzhou 310056, China (e-mail: matengfei.zjut@gmail.com, vencent\_wang@outlook.com, dongweixu@zjut.edu.cn, xuanqi@zjut.edu.cn).}
\thanks{Zhuangzhi Chen is with the Institute of Cyberspace Security, Zhejiang University of Technology, Hangzhou 310023, China, and also with the Research Center of Electromagnetic SpaceSecurity, Binjiang Institute of Artificial Intelligence, ZJUT, Hangzhou, 310056, China (e-mail: zzch@zjut.edu.cn)}
\thanks{Yun Lin is with the College of Information and Communication Engineering, Harbin Engineering University, Harbin, China (e-mail: linyun@hrbeu.edu.cn).}
\thanks{Guan Gui is with the College of Telecommunications and Information Engineering, Nanjing University of Posts and Telecommunications, Nanjing 210003, China (e-mail: guiguan@njupt.edu.cn).}}

\markboth{Journal of \LaTeX\ Class Files,~Vol.~14, No.~8, August~2021}%
{Shell \MakeLowercase{\textit{et al.}}: A Sample Article Using IEEEtran.cls for IEEE Journals}


\maketitle

\begin{abstract}
With the rapid development of wireless communications and the growing complexity of digital modulation schemes, traditional manual modulation recognition methods struggle to extract reliable signal features and meet real-time requirements in modern scenarios. Recently, deep learning based Automatic Modulation Recognition (AMR) approaches have greatly improved classification accuracy. However, their large model sizes and high computational demands hinder deployment on resource-constrained devices. Model pruning provides a general approach to reduce model complexity, but existing weight, channel, and layer pruning techniques each present a trade-off between compression rate, hardware acceleration, and accuracy preservation. To this end, in this paper, we introduce FCOS, a novel \textbf{F}ine-to-\textbf{CO}arse two-\textbf{S}tage pruning framework that combines channel-level pruning with layer-level collapse diagnosis to achieve extreme compression, high performance and efficient inference. In the first stage of FCOS, hierarchical clustering and parameter fusion are applied to channel weights to achieve channel-level pruning. Then a \textbf{La}yer \textbf{C}ollapse \textbf{D}iagnosis (LaCD) module uses linear probing to identify layer collapse and removes the collapsed layers due to high channel compression ratio. Experiments on multiple AMR benchmarks demonstrate that FCOS outperforms existing channel and layer pruning methods. Specifically, FCOS achieves 95.51\% FLOPs reduction and 95.31\% parameter reduction while still maintaining performance close to the original ResNet56, with only a 0.46\% drop in accuracy on Sig2019-12. Code is available at \url{https://github.com/yaolu-zjut/FCOS}.
\end{abstract}


\section{Introduction}

\IEEEPARstart{W}{ith} the rapid development of wireless communication technology, especially the complexity and diversification of modulation technology in digital communication systems, traditional manual modulation recognition methods have been unable to cope with modern communication needs. Specifically, modulated signals are often encrypted, source coded, and channel coded before transmission, which significantly increases the difficulty of signal feature extraction~\cite{walenczykowska2016type,li2019wavelet,triantafyllakis2017phasma,vuvcic2017cyclic,abdelmutalab2016automatic}. Besides, the diversity of modulation methods makes manual recognition inefficient and error-prone. In this context, Automatic Modulation Recognition (AMR) uses deep learning algorithm~\cite{o2018over,o2016convolutional,lin2020improved,chen2021signet,lin2020contour,tu2020complex,lin2020adversarial,zhang2023lightweight,hou2024multi,xu2025mclrl} to realize intelligent judgment of modulation types, which can not only improve the accuracy and real-time performance of signal analysis in complex electromagnetic environments, but also provide core technical support for scenarios such as cognitive radio~\cite{wang2019data}, spectrum sensing~\cite{lim2012automatic}, signal surveillance~\cite{boutte2009feature}, and interference identification~\cite{zhang2024automatic}. Therefore, in-depth research on it is of great significance to promote the intelligent development of communication systems.

In recent years, many deep learning based methods have been proposed specifically for AMR. For instance, inspired by the VGG~\cite{simonyan2014very} architecture in image classification, O'Shea et al.~\cite{o2018over} design a 1D-CNN model for short radio signal classification. Subsequently, they further introduce a slim 2D-CNN that processes both in-phase (I) and quadrature (Q) signal sequences in the time domain for radio modulation recognition~\cite{o2016convolutional}. More recently, Chen et al.~\cite{chen2021signet} present the S2M operator, a sliding square mechanism that converts input signals into square feature matrices, thereby enabling the use of more advanced classification models. To improve the classification accuracy in low signal-to-noise ratio (SNR) environments, Zhang et al.~\cite{zhang2023amc} propose a novel AMC-Net that denoises the input signal in the frequency domain and performs multi-scale and effective feature extraction.

Although these deep learning based AMR methods have achieved impressive results in signal classification tasks, the models these methods used exhibit high computational complexity and large model sizes. This leads to slow inference speeds, high bandwidth requirements, and heavy resource consumption, making them hard to use on everyday smart devices and real-world communication systems with limited resources~\cite{zeng2024efficient,jameel2024towards,xu2023edge,shi2023task}. To this end, researchers begin to focus on designing lightweight AMR models and propose a variety of optimization strategies. Among these strategies, model pruning~\cite{zhang2018systematic,ma2020pconv,aghli2021combining,meng2020pruning,molchanov2019importance,he2017channel,zhuang2018discrimination,lu2022understanding,tang2023sr,chen2018shallowing,huang2018data,lu2024generic,lu2024reassessing} has received extensive research attention due to its algorithmic universality. 

Specifically, model pruning can be roughly divided into three categories: weight pruning~\cite{zhang2018systematic,ma2020pconv}, channel pruning~\cite{meng2020pruning,chen2023rgp} and layer pruning~\cite{lu2022understanding,tang2023sr}. As shown in \cref{tab:comparsion}, all three methods have their own shortcomings. Weight pruning can achieve extremely high compression rates, but it requires specialized hardware to achieve effective acceleration, and the inference acceleration effect in actual deployment is not obvious. By contrast, channel pruning and layer pruning can achieve effective inference acceleration on ordinary computing devices, but their model compression capabilities are relatively limited. Specifically, channel pruning must ensure that a certain number of channels are left in each layer for effective reasoning, which makes it impossible to achieve extremely high compression rates. Layer pruning is inherently coarse-grained, as it removes entire layers. To preserve model accuracy, only a limited number of layers can be pruned, which makes achieving a higher compression rate challenging.

In view of these, we want to design a new pruning method that can balance high compression rate and efficient inference acceleration. We choose channel pruning as the basic solution because it has a finer pruning granularity than layer pruning, which can achieve a better balance in compression rate and performance. Through analysis, we find that the reason why channel pruning cannot achieve a high compression rate is that excessive pruning will cause the model structure to collapse, feature extraction to fail, and ultimately result in a sharp drop in model performance (see \cref{tab:Layer Collapse}). Therefore, we design a \textbf{F}ine-to-\textbf{CO}arse two-\textbf{S}tage pruning method \textbf{FCOS}. In the first stage, FCOS implements a standardized channel pruning procedure. It utilizes a hierarchical clustering algorithm to cluster the weights of all channels in each layer, and then uses the model merging technique to fuse the parameters within the same cluster. In the second stage, FCOS introduces a \textbf{La}yer \textbf{C}ollapse \textbf{D}iagnosis method (LaCD), which utilizes linear probing to diagnose layer collapse and removes collapsed layers. With the support of FCOS, we have successfully overcome the model collapse problem that traditional channel pruning is prone to at high compression rates, achieving high compression rates while maintaining high accuracy and significant inference acceleration. This improvement provides solid technical support for the efficient deployment of deep learning based AMR models in resource-constrained scenarios.

\begin{table*}[htbp]
  \centering
  \caption{Comparison of the proposed method FCOS with existing different types of pruning methods.}
    \begin{tabular}{ccccc}
    \toprule
    Type  & No specialized hardware required & Not constrained by model depth & Very high compression ratio & Obvious acceleration effect \\
    \midrule
    Weight Pruning &   \XSolid  & \Checkmark   & \Checkmark   & \XSolid \\
    Channel Pruning & \Checkmark   & \XSolid    & \XSolid    & \Checkmark \\
    Layer Pruning & \Checkmark   & \Checkmark   & \XSolid    & \Checkmark \\
    FCOS   & \Checkmark   & \Checkmark   & \Checkmark   & \Checkmark \\
    \bottomrule
    \end{tabular}%
  \label{tab:comparsion}%
\end{table*}%

To evaluate the efficacy of FCOS, we conduct comprehensive experiments across three widely-used benchmark datasets (RML2016.10a~\cite{o2016radio}, Sig2019-12~\cite{chen2021signet}, and RML2018.01a~\cite{o2018over}) and three representative AMR models (CNN1D, ResNet50, and SigNet50). Experimental results show that compared with the state-of-the-art channel pruning techniques (RFP~\cite{shao2021filter}, FPGM~\cite{he2019filter}, L1-norm~\cite{li2016pruning}, SFP~\cite{he2018soft}, BNP~\cite{chen2023channel}) and layer pruning methods (random, LCP~\cite{chen2018shallowing, wang2019dbp}, SR-init~\cite{tang2023sr}, PSR~\cite{lu2024generic}), FCOS achieves superior performance while achieving extremely high compression rate. For example, FCOS achieves 95.51\% FLOPs reduction and 95.31\% parameter reduction while still maintaining performance close to the original ResNet56, with only a 0.46\% drop in accuracy on Sig2019-12.

To summarize, our main contributions are three-fold:
\begin{itemize}
\item[$\bullet$] We introduce a fine-to-coarse two-stage pruning method FCOS, which maintains high accuracy and significant inference acceleration while ensuring extremely high compression rate. In the first stage, FCOS uses hierarchical clustering to cluster the weights of all channels of each convolutional layer, and then uses the model merging technique to fuse the parameters within the same cluster to achieve pruning in the width dimension.
\item[$\bullet$] However, when performing channel pruning, the performance of the pruned model will collapse due to excessive channel pruning rate. To solve this problem, we introduce a \textbf{La}yer \textbf{C}ollapse \textbf{D} diagnosis method (LaCD) in the second stage of FCOS, which uses linear probing to diagnose layer collapse and removes the collapsed layers to obtain the optimal model.
\item[$\bullet$] Extensive experiments demonstrate that FCOS outperforms $5$ state-of-the-art channel pruning approaches and $4$ layer pruning baselines across $3$ benchmark datasets (RML2016.10a, Sig2019-12, RML2018.01a) and $3$ AMR models (CNN1D, ResNet50, SigNet50) while achieving extremely high model compression rate.
\end{itemize}

In the remainder of this paper, we first introduce related works on deep learning based lightweight AMR, model fusion and model pruning in \cref{sec:Related Works}. In \cref{sec:model purify}, we delve into the fine-to-coarse two-stage pruning method \textbf{FCOS}. Then experiments are discussed in \cref{sec:Experiments}. Finally, the paper concludes in \cref{sec:Conclusion}.

\section{Related Works}
\label{sec:Related Works}
\textbf{Deep Learning based Lightweight AMR.} While deep learning based AMR models have demonstrated promising results, their practical deployment is constrained by substantial resource requirements arising from their parameter-heavy architectures and computationally intensive operation. To address this challenge, researchers have proposed various lightweighting methods~\cite{fu2021lightweight,zhang2022amr,chen2023channel,lin2020improved,guo2024ultra}. For example, Zhang et al.~\cite{zhang2022amr} develop an AMR model based on neural architecture search (NAS) to dynamically optimize model structure and parameters through automated search. Subsequently, Zhang et al~\cite{zhang2023lightweight} further propose a progressively differentiable architecture search method. Besides, Chen et al.~\cite{chen2023channel} use the $\gamma$ coefficient of batch normalization layer to evaluate channel importance and guide model pruning. Lin et al.~\cite{lin2020improved} judge the importance of channels based on activation maximization. Recently, Lu et al~\cite{lu2024generic} propose a three-stage layer pruning method based on Fisher optimal segmentation and a training-free performance estimation method SynFlow~\cite{tanaka2020pruning}. Although these methods have achieved good results, they also have some limitations. For example, the typical NAS~\cite{zhang2022amr,zhang2023lightweight} process requires hundreds to thousands of GPU training days, which not only leads to high economic costs, but also causes a lot of carbon emissions. Whereas pruning techniques~\cite{chen2023channel,lin2020improved,lu2024generic} can avoid extensive architecture search time, they struggle to achieve significant model compression rates while maintaining accuracy, which is a key requirement for deploying AMR models on resource-constrained edge devices.

\textbf{Model fusion} (also known as model merging) is an emerging technique that merges the parameters of multiple deep learning models into a single one, combining the abilities of different models to make up for the biases and errors of a single model to achieve better performance~\cite{li2023deep}. The simplest model fusion is linear interpolation. In addition to simple linear interpolation, Matena et al.~\cite{matena2022merging} propose Fisher Merging, which uses the Fisher information of the model as the posterior precision matrix to perform a Laplacian approximation. Jin et al~\cite{jin2022dataless} introduce RegMean to minimize the $l_{2}$ distance between the merged model and other multiple models trained on various tasks. Recently, PAPA~\cite{jolicoeur2023population} trains multiple models with the same initialization but slightly varied data configurations—such as different orderings, augmentations, and regularizations, average these models every few epochs to improve the generalization of the model. Furthermore, Rame et al~\cite{rame2023model} introduce model ratatouille to recycle the multiple fine-tunings of the same foundation model on diverse auxiliary tasks. In this paper, due to the high redundancy of model parameters, we employ model fusion to combine similar weights, thereby achieving channel slimming.

\textbf{Model pruning} is a widely recognized model compression technique, which can be broadly categorized into weight pruning~\cite{zhang2018systematic,ma2020pconv,aghli2021combining}, channel pruning~\cite{meng2020pruning,molchanov2019importance,he2017channel,zhuang2018discrimination}, and layer pruning~\cite{lu2022understanding,tang2023sr,chen2018shallowing,huang2018data,lu2024generic,lu2024reassessing}, depending on the level of granularity. Specifically, while weight pruning can achieve a high compression ratio by zeroing out certain weights, its applicability is largely restricted to specialized software~\cite{park2016faster} or hardware~\cite{han2016eie} devices. In contrast, channel pruning and layer pruning have no such constraints. Channel pruning improves inference speed by removing specific channels from each layer. For instance, RFP~\cite{shao2021filter} selects important channels based on the information entropy of feature maps, while FPGM~\cite{he2019filter} uses the geometric median~\cite{fletcher2008robust}. SFP~\cite{he2018soft} adopts a dynamic, soft pruning strategy based on the ${\ell _2}$-norm, whereas BNP~\cite{chen2023channel} determines channel importance through the $\gamma$ scaling factor of batch normalization layers. Layer pruning selects those layers that contribute minimally to performance for removal. For example, Elkerdawy et al.~\cite{elkerdawy2020filter} use existing channel criteria to calculate a per-layer importance score in one-shot. Chen et al.~\cite{chen2018shallowing} use Linear Classifier Probes (LCPs) to examine how each layer boosts performance and eliminate layers that contribute little. Lu et al.~\cite{lu2022understanding} propose a training-free metric called modularity to evaluate the class separation in intermediate representations, pruning layers that display zero or negative growth in modularity. Lu et al.~\cite{lu2024generic} propose a three-stage hierarchical pruning framework that first decomposes the model into semantic blocks, prunes redundant layers via contribution-based layer importance criteria, and finally reassembles and fine-tunes the compact model. Recently, Lu et al.~\cite{lu2024reassessing} find through a large number of experiments that pruning the last several layers performs better than many complex pruning metrics.

Although channel pruning and layer pruning are not tied to specific devices, each has its own limitations. In particular, the compression rate of channel pruning is restricted by the model's depth, whereas layer pruning cannot achieve high compression rates due to its coarse granularity. To address this limitation, this paper introduces a two-stage pruning method from fine to coarse, which integrates the advantages of channel pruning and layer pruning, overcoming the low compression ceiling of individual pruning strategies.

\section{Method}
\label{sec:model purify}
\begin{figure*}[t]
	\centering
    \includegraphics[width=0.99\textwidth]{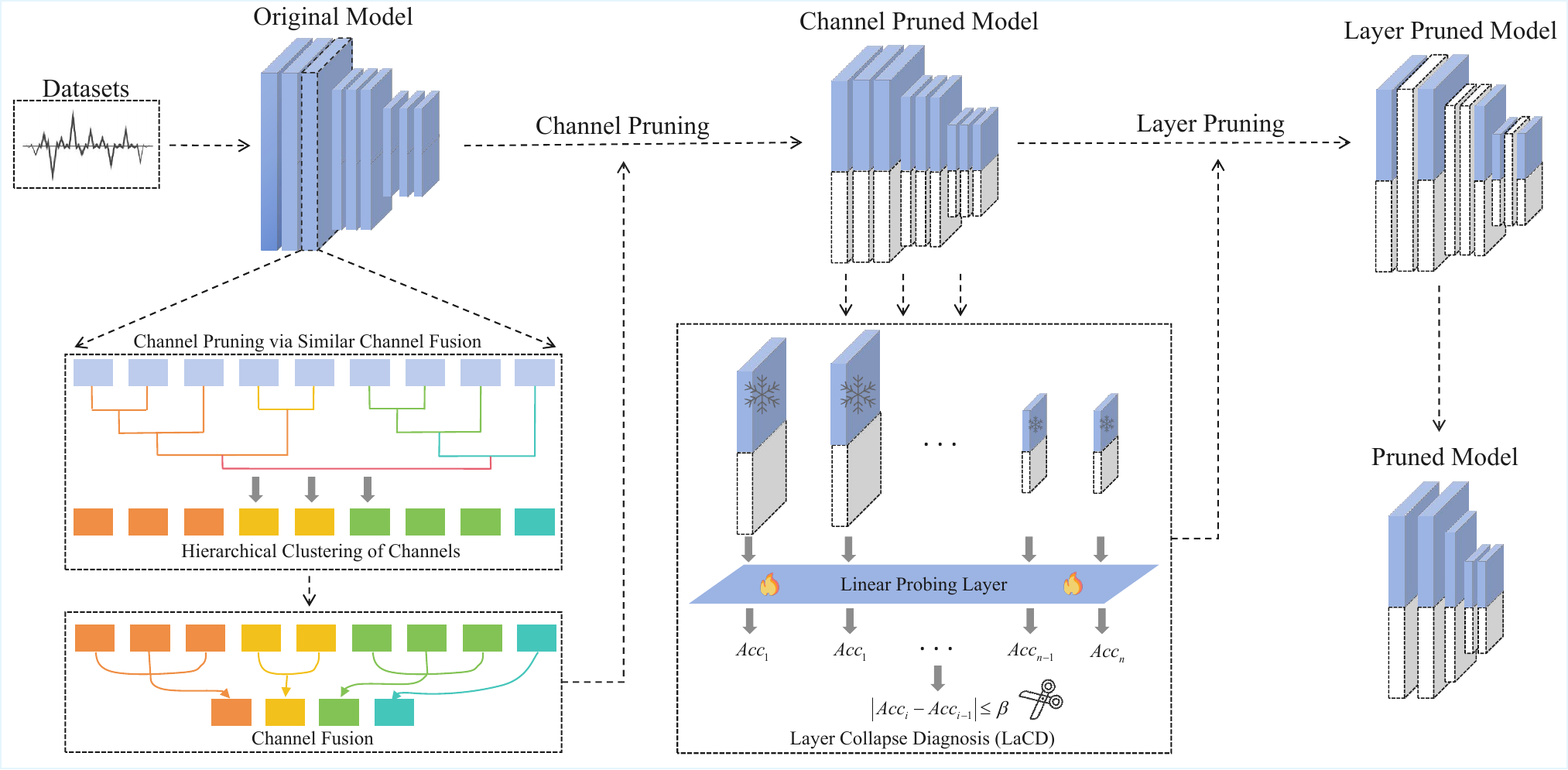} 
 \caption{The pipeline of FCOS. In the first stage, hierarchical clustering and parameter fusion are applied to channel weights to achieve channel-level pruning. Then a layer collapse diagnosis (LaCD) module uses linear probing to identify layer collapse and removes the collapsed layers due to high channel compression ratio.} 
 \label{fig:pipeline}
\end{figure*}

In this section, we delve into our fine-to-coarse two-stage pruning method \textbf{FCOS}. In the first stage, FCOS uses a hierarchical clustering algorithm to cluster the weights of all channels in each convolutional layer, and then uses the model merging technique to fuse the parameters within the same cluster, achieving pruning in the width dimension. However, when the channel compression rate is too high, the performance of the pruned model will collapse due to excessive pruning. To address this issue, in the second stage of FCOS, we introduce a \textbf{La}yer \textbf{C}ollapse \textbf{D}iagnosis method (LaCD), which utilizes linear probing to diagnose layer collapse and removes collapsed layers to obtain an optimal model. We provide the overall process of FCOS in \cref{fig:pipeline}.

\subsection{Channel Pruning via Similar Channel Fusion}
\label{sec:Channel Pruning via Similar Channel Fusion}
Given a pre-trained model $\mathcal{M}$ with $L$ convolutional layers, the weight tensor of $i$-th convolutional layer can be represented as $w_i \in \mathbb{R}^{c^{out}_{i} \times c^{in}_{i} \times h_{i} \times l_{i}}$ for $i \in \{1,2,\cdots, L\}$. $c^{out}_{i}$ and $c^{in}_{i}$ denote the number of input channels and output channels of the $i$-th convolutional layer, while $h_i$ and $l_i$ are the width and height of the convolution kernel. For each output channel of the $i$-th convolutional layer, we vectorize its weight tensor as $\operatorname{vec}\left(w_{i,j}\right) \in \mathbb{R}^{d_{i}}$, where $d_{i}=c^{in}_{i} \times h_i \times l_i$ and $j \in \{1,2,\cdots, c^{out}_{i}\}$. Then, we use cosine similarity (\cref{eq:cos}) to calculate similarity between any two output channels.
\begin{equation}
\operatorname{sim}\left(w_{i,j}, w_{i,k}\right)=\frac{\left\langle\operatorname{vec}\left(w_{i,j}\right), \operatorname{vec}\left(w_{i,k}\right)\right\rangle}{\left\|\operatorname{vec}\left(w_{i,j}\right)\right\|\left\|\operatorname{vec}\left(w_{i,k}\right)\right\|}.
\label{eq:cos}
\end{equation}
The corresponding distance is then derived as:
\begin{equation}
D_{jk}^i = 1-\operatorname{sim}\left(w_{i,j}, w_{i,k}\right).
\label{eq:dis}
\end{equation}
It is worth noting that, in \cref{sec: Ablation Study}, we demonstrate that cosine similarity can be replaced by any similarity metric. After obtaining the distance matrix $D_{jk}^i$, we employ hierarchical clustering algorithm~\cite{johnson1967hierarchical} to divide channels into $n$ groups, where 
\begin{equation}
n = \lfloor|c^{out}_{i}|\times\epsilon\rfloor.
\label{eq:pruing rate}
\end{equation}
Herein, $\epsilon$ is a predefined channel pruning rate. Specifically, we first treat each channel as a separate cluster, calculate the distance between all clusters, and gradually merge the two clusters with the smallest distance. There are three methods to calculate the distance between two clusters: single linkage, complete linkage, and average linkage. In this paper, we choose average linkage and implement it with SciPy\footnote{https://scipy.org/}:
\begin{equation}
\mathcal{D}_i\left(c_a, c_b\right)=\frac{1}{\left|c_a\right|\left|c_b\right|} \sum_{j \in c_a} \sum_{k \in c_b}D_{jk}^i,
\end{equation}
where $c_a$ and $c_b$ are two clusters in $i$-th convolutional layer. A key reason for clustering based on the weight similarity of the channel is that the learned weights inherit the capabilities of the pre-trained model~\cite{matena2022merging}. By comparing the weight tensors directly, we can detect channels whose parameters exhibit highly similar patterns and prune them with minimal cost.

\begin{algorithm}[t]
    \caption{Channel Pruning via Similar Channel Fusion}
    \label{algorithm:Channel Pruning via Similar Channel Fusion}
    \textbf{Input}: A pre-trained model $\mathcal{M}$, Pruning rate $\epsilon$.\\
    \textbf{Output}: A pruned model $\mathcal{M}^{*}$. \\
    \begin{algorithmic}[1] 
    \For{$i = 1$ to $L$}
        \For{$j = 1$ to $c^{out}_{i}$} \label{li:loop_begin}
        \For{$k = 1$ to $c^{out}_{i}$}
        \State $\operatorname{sim}\left(w_{i,j}, w_{i,k}\right)=\frac{\left\langle\operatorname{vec}\left(w_{i,j}\right), \operatorname{vec}\left(w_{i,k}\right)\right\rangle}{\left\|\operatorname{vec}\left(w_{i,j}\right)\right\|\left\|\operatorname{vec}\left(w_{i,k}\right)\right\|}.$
        \State $D_{jk}^i = 1-\operatorname{sim}\left(w_{i,j}, w_{i,k}\right)$
        \EndFor
        \EndFor
        \State \text{Employ hierarchical clustering.} 
        \State \text{Divide $c^{out}_{i}$ channels into $\lfloor|c^{out}_{i}|\times\epsilon\rfloor$ groups.}
        \For{channels in each group}
        \State Weight fusion via \cref{eq:fusion}.
        \EndFor \label{li:loop_end}
    \If{$c^{in}_{i}$ also needs to be pruned}
    \State Repeat \cref{li:loop_begin}-\cref{li:loop_end} to prune input channels $c^{in}_{i}$.
    \EndIf
    \EndFor
    \end{algorithmic}
\end{algorithm}

After hierarchical clustering, the weight tensors within the same group have high similarity, indicating that there is information redundancy. To this end, we utilize model fusion technology~\cite{li2023deep,huang2024emr,yadav2023ties} to fuse the parameters of channels in the same cluster to generate a representative channel. For the $|c_a|$ channels contained in the cluster $c_a$, its fusion weight can be obtained by simply averaging: 
\begin{equation}
w_{i,c_a}^*=\frac{1}{|c_a|} \sum_{j=1}^{|c_a|} w_{i,j}.
\label{eq:fusion}
\end{equation}
After channel fusion, the number of output channels of $i$-th layer is reduced from $c^{out}_{i}$ to $n$, yielding model slimming. Besides, the fusion operation helps to reduce the adverse effects caused by noise or anomalies in individual channels, thereby improving the generalization~\cite{izmailov2018averaging,wortsman2022model} and robustness~\cite{gao2022revisiting,croce2023seasoning} of the pruned model.

It is worth noting that the output channel of the previous convolutional layer of the pre-trained model is associated with the input channel of the next convolutional layer. This means that, except for the first layer, the input and output channels of other layers need to be pruned to achieve inter-layer parameter matching. To this end, we repeat the similar channel fusion operator on the input channels $c^{in}_{i}$. In \cref{sec: Ablation Study}, our experiments confirm that the sequential ordering of weight fusion operations - whether applied to input or output channels first - demonstrates negligible impact on the performance of the pruned model. Finally, we can obtain the slimming model $\mathcal{M}^{*}$. We provide a summary of this pipeline in \cref{algorithm:Channel Pruning via Similar Channel Fusion}.

\subsection{Layer Collapse Diagnosis via Linear Probing}
\label{sec:Layer Collapse Diagnosis via Linear Probing}
In \cref{sec:Channel Pruning via Similar Channel Fusion}, we have obtained the pruned model using \cref{algorithm:Channel Pruning via Similar Channel Fusion}. Next, the most intuitive method is to directly fine-tune the pruned model. However, we find through experiments that when the channel pruning rate is high, the performance of the pruned model may collapse. Specifically, as shown in \cref{tab:Layer Collapse}, when the pruning rate reaches 99\%, there is a probability that the model performance will collapse. Notably, this collapse is not simply a result of performance loss from high compression, but rather due to the breakdown of the model's structure induced by such extreme compression.

\begin{table}[t]
  \centering
  \caption{When the pruning rate is too high, there is a probability that the model performance will collapse. PR DENOTES THE PRUNING RATE. The original model accuracy is $0.6234$.}
  \resizebox{0.49\textwidth}{!}{
    \begin{tabular}{cccccc}
    \toprule
    Pruning Model Status & Original Params & Original FLOPs & Params PR & FLOPs PR & Acc \\
    \midrule
    Normal & 852.79K & 42.23M & 99.10\% & 99.14\% & 0.5545 \\
    Abnormal & 852.79K & 42.23M & 99.10\% & 99.14\% & 0.4652 \\
    \bottomrule
    \end{tabular}}
  \label{tab:Layer Collapse}%
\end{table}%

To address this challenge, we introduce a \textbf{La}yer \textbf{C}ollapse \textbf{D}iagnosis method (LaCD), which utilizes linear probing to detect layer collapse in pruned models due to excessive channel pruning. Specifically, given a pre-pruned model $\mathcal{M}^{*}$ obtained by \cref{algorithm:Channel Pruning via Similar Channel Fusion} and a labeled dataset $\mathcal{D}ata$, we first fine-tune the model for several epochs (e.g., $20$ epochs). Fine-tuning with a small number of epochs can quickly restore the pruned model to a certain level of accuracy, thereby providing a stable feature distribution for subsequent collapse layer diagnosis. This short-cycle fine-tuning can significantly reduce the overall computing cost and avoid wasting too many resources in the early stages. Then, we initialize the set of layers to be pruned $\mathcal{P}_r= \{\}$ and perform layer collapse diagnosis on each suspicious layer in the model (such as convolutional layers or fully connected layers). We freeze the gradients of $\mathcal{M}^{*}$ at this stage to avoid distribution drift caused by parameter updates. For each candidate layer $i$, we extract the feature representation $F_i$ of the layer output from the train set $\mathcal{D}ata = \{X,Y\}$:
\begin{equation}
F_i=f_i \circ \cdots \circ f_2 \circ f_1(X),
\end{equation}
where $f_i$ represents the function of the $i$-th layer of the model $\mathcal{M}^{*}$, $\circ$ indicates function composition and $X$ denotes the input data. After obtaining the feature representation $F_i$, we train a linear classifier $lc$ using \cref{eq:classification} with $\{(F_i,Y)\}$ and get the classification accuracy $Acc_i$ on the test set.
\begin{equation}
    \mathop{min}\limits_{\theta} \frac{1}{|Y|} \sum_{j=1}^{|Y|}{\mathcal{L}(lc(F_{i,j}), y_j)},
  \label{eq:classification}
\end{equation} 
where $\mathcal{L}$, $\theta$ and $y_j \in Y$ denote the loss function, the parameters of the linear classifier and the ground-truth label, respectively. By analyzing the layer-level features in the frozen state, the layers that have little impact on the overall performance can be identified at a lower computational cost and incorporated into $\mathcal{P}_r$. Here, we define the impact of the $i$-th layer on the model performance as follows:
\begin{equation}
|Acc_i-Acc_{i-1}| \leq \beta,
\end{equation}
where $\beta$ is a predefined layer redundancy threshold. Finally, we remove collapsed layers in $\mathcal{P}_r$, fine-tune the model for $e$ epochs and obtain an optimal model $\mathcal{M}^{**}$. The overall pipeline is summarized in \cref{algorithm:linear probe}.

\begin{table*}[t]
  \centering
  \caption{Introduction to baseline methodologies.}
  \resizebox{0.99\textwidth}{!}{
    \begin{tabular}{cccc}
    \toprule
    Method & Type & Pruning Type & \multicolumn{1}{c}{Description} \\
    \midrule
    RFP   & channel pruning & hard pruning & \multicolumn{1}{c}{RFP~\cite{shao2021filter} selects important channels based on the information entropy of feature maps.} \\
    FPGM  & channel pruning & soft pruning &\multicolumn{1}{c}{FPGM~\cite{he2019filter} prunes redundant channels based on the geometric median of the channels.} \\
    L1-norm & channel pruning & hard pruning &\multicolumn{1}{c}{L1-norm~\cite{li2016pruning} measures the relative importance of a channel in each layer by calculating its $\ell_1$-norm.} \\
    SFP   & channel pruning & soft pruning &\multicolumn{1}{c}{SFP~\cite{he2018soft} dynamically prunes the channels using $\ell_2$-norm in a soft manner.} \\
    BNP   & channel pruning & hard pruning &\multicolumn{1}{c}{BNP~\cite{chen2023channel} measures the channel importance using the $\gamma$ scale factor of the batch normalization layer.} \\
    \midrule
    Random & layer pruning &hard pruning & \multicolumn{1}{c}{Layers are removed based on a random selection rather than through careful analysis.} \\
    LCP   & layer pruning & hard pruning &\multicolumn{1}{c}{LCP~\cite{chen2018shallowing,wang2019dbp} uses linear classifier probes to delete unimportant layers.} \\
    SR-init & layer pruning &hard pruning & \multicolumn{1}{c}{SR-init~\cite{tang2023sr} prunes layers that are not sensitive to stochastic re-initialization.} \\
    PSR   & layer pruning & hard pruning &PSR~\cite{lu2024generic} decomposes the model into consecutive blocks and identifies layers that need to be preserved within each block with SynFlow. \\
    \bottomrule
    \end{tabular}}
  \label{tab:baseline intro}%
\end{table*}%

\begin{algorithm}[t]
    \caption{Layer Collapse Diagnosis (LaCD) Method}
    \label{algorithm:linear probe}
    \textbf{Input}: A pruned model $\mathcal{M}^{*}$ obtained by \cref{algorithm:Channel Pruning via Similar Channel Fusion}, a labeled train set $\mathcal{D}ata = \{X,Y\}$, a layer redundancy threshold $\beta$, fine-tuning epochs $e$.\\
    \textbf{Output}: A further pruned model $\mathcal{M}^{**}$. \\
    \begin{algorithmic}[1] 
    \State Initialize the set of layers to be pruned $\mathcal{P}_r= \{\}$.
    \State Fine-tune the pruned model $\mathcal{M}^{*}$ for $20$ epochs.
    \State Freeze the gradients of $\mathcal{M}^{*}$.
    \For{$i = 1$ to $L$}
    \State Extract feature representation $F_i$ from layer $i$ using $X$.
    \State Train a linear classifier $lc$ on $F_i$.
    \State Compute classification accuracy $Acc_i$.
    \If{$|Acc_i-Acc_{i-1}| \leq \beta$}
        \State Append $i$ into $\mathcal{P}_r$.
    \EndIf
    \EndFor
    \State Remove collapsed layers in $\mathcal{P}_r$.
    \State Fine-tune the model for $e$ epochs.
    \end{algorithmic}
\end{algorithm}




\section{Experiments}
\label{sec:Experiments}
In this section, we verify the effectiveness of FCOS on $4$ models across $3$ datasets.

\begin{table*}[htbp]
  \centering
  \caption{PRUNING RESULTS OF CNN1D AND ResNet56 ON RML2016.10a, RML2018.01a and Sig2019-12. PR DENOTES THE PRUNING RATE.}
  \resizebox{0.9\textwidth}{!}{
    \begin{tabular}{c|c|ccccccccc}
    \toprule
    Model & Dataset & Method & Pruning Type & Original Acc(\%) & Acc(\%) & $\triangle$Acc(\%) & $\triangle$FLOPs & $\triangle$Params & FLOPs PR & Params PR \\
    \midrule
    \multirow{30}[6]{*}{ResNet56} & \multirow{10}[2]{*}{RML2016.10a} & RFP   & Channel & 60.70  & 54.72 & -5.98 ↓ & -33.74M & -749.19K & 80.50\% & 87.90\% \\
          &       & FPGM  & Channel & 62.61 & 59.43 & -3.18 ↓ & - & -765.76K & - & 89.80\% \\
          &       & L1-norm & Channel & 62.56 & 60.70  & -1.86 ↓& -36.51M & -743.65K & 87.10\% & 87.20\% \\
          &       & SFP   & Channel & 60.70  & 58.72 & -1.98 ↓& - & -381.94K & - & 44.80\% \\
          &       & BNP   & Channel & 62.63 & 61.39 & -1.24 ↓ & -36.51M & -743.65K & 87.10\% & 87.20\% \\
          &       & Random & Layer & 62.06 & 60.08 & -1.98 ↓ & -35.64M & -703.74K & 85.00\% & 82.50\% \\
          &       & LCP   & Layer & 62.06 & 60.84 & -1.22 ↓ & -36.81M & -773.05K & 87.80\% & 90.60\% \\
          &       & SR-init & Layer & 62.06 & 61.02 & -1.04 ↓ & -35.64M & -703.74K & 85.00\% & 82.50\% \\
          &       & PSR   & Layer & 62.06 & 61.10  & -0.96 ↓ & -36.82M & -759.16K & 87.80\% & 89.00\% \\
          &       & \cellcolor[rgb]{ .839,  .863,  .894}FCOS & \cellcolor[rgb]{ .839,  .863,  .894}Channel+Layer & \cellcolor[rgb]{ .839,  .863,  .894}62.06 & \cellcolor[rgb]{ .839,  .863,  .894}62.12 & \cellcolor[rgb]{ .839,  .863,  .894}+0.06 ↑ & \cellcolor[rgb]{ .839,  .863,  .894}-37.33M & \cellcolor[rgb]{ .839,  .863,  .894}-753.38K & \cellcolor[rgb]{ .839,  .863,  .894}88.39\% & \cellcolor[rgb]{ .839,  .863,  .894}88.34\% \\
\cmidrule{2-11}          & \multirow{10}[2]{*}{Sig2019-12} & RFP   & Channel & 66.20  & 55.21 & -10.99 ↓ & -161.57M & -765.60K & 96.40\% & 89.80\% \\
          &       & FPGM  & Channel & 66.22 & 50.23 & -15.99 ↓ & - & -778.81K & - & 91.30\% \\
          &       & L1-norm & Channel & 67.16 & 64.13 & -3.03 ↓ & -163.60M & -761.35K & 97.60\% & 89.30\% \\
          &       & SFP   & Channel & 66.22 & 17.22 & -49.00 ↓ & - & -483.90K & - & 56.70\% \\
          &       & BNP   & Channel & 67.16 & 61.53 & -5.63 ↓ & -163.60M & -761.35K & 97.60\% & 89.30\% \\
          &       & Random & Layer &  67.91 & 61.77 & -6.14 ↓ & -152.04M & -777.73K & 90.70\% & 91.20\% \\
          &       & LCP   & Layer &  67.91 & 64.33 & -3.58 ↓ & -152.04M & -777.73K & 90.70\% & 91.20\% \\
          &       & SR-init & Layer &  67.91 & 64.90  &  -3.01 ↓ & -152.04M & -777.73K & 90.70\% & 91.20\% \\
          &       & PSR   & Layer & 67.91 & 64.93 & -2.98 ↓ & -152.04M & -777.73K & 90.70\% & 91.20\% \\
          &       & \cellcolor[rgb]{ .839,  .863,  .894}FCOS & \cellcolor[rgb]{ .839,  .863,  .894}Channel+Layer & \cellcolor[rgb]{ .839,  .863,  .894}67.91 & \cellcolor[rgb]{ .839,  .863,  .894}67.45 & \cellcolor[rgb]{ .839,  .863,  .894}-0.46 ↓ & \cellcolor[rgb]{ .839,  .863,  .894}-161.32M & \cellcolor[rgb]{ .839,  .863,  .894}-812.83K & \cellcolor[rgb]{ .839,  .863,  .894}95.51\% & \cellcolor[rgb]{ .839,  .863,  .894}95.31\% \\
\cmidrule{2-11}          & \multirow{10}[2]{*}{RML2018.01a} & RFP   & Channel &  92.00 & 42.72 & -49.28 ↓ & -328.70M & -768.86K & 98.00\% & 90.20\% \\
          &       & FPGM  & Channel &  91.95 & 79.56 & -12.39 ↓ & - & -778.03K & - & 91.20\% \\
          &       & L1-norm & Channel &  92.56 & 77.95 & -14.61 ↓ & -331.28M & -760.57K & 98.80\% & 89.20\% \\
          &       & SFP   & Channel &  91.95 & 32.75 & -59.20 ↓ & - & -483.12K & - & 56.70\% \\
          &       & BNP   & Channel & 92.00    & 82.79 &  -9.21 ↓ & -331.28M & -760.57K & 98.80\% & 89.20\% \\
          &       & Random & Layer &  89.70 & 83.45 & -6.25 ↓ & -304.09M & -777.73K & 90.70\% & 91.10\% \\
          &       & LCP   & Layer &  89.70 & 82.70  &  -7.00 ↓ & -304.09M & -777.73K & 90.70\% & 91.10\% \\
          &       & SR-init & Layer &  89.70 & 82.72 &  -6.98 ↓ & -304.09M & -777.73K & 90.70\% & 91.10\% \\
          &       & PSR   & Layer &  89.70 & 84.07 &  -5.63 ↓ & -304.09M & -777.73K & 90.70\% & 91.10\% \\
          &       & \cellcolor[rgb]{ .839,  .863,  .894}FCOS & \cellcolor[rgb]{ .839,  .863,  .894}Channel+Layer & \cellcolor[rgb]{ .839,  .863,  .894}89.70 & \cellcolor[rgb]{ .839,  .863,  .894}90.61 & \cellcolor[rgb]{ .839,  .863,  .894}+0.91 ↑ & \cellcolor[rgb]{ .839,  .863,  .894}-304.33M & \cellcolor[rgb]{ .839,  .863,  .894}-768.55K & \cellcolor[rgb]{ .839,  .863,  .894}90.09\% & \cellcolor[rgb]{ .839,  .863,  .894}90.03\% \\
    \midrule
    \multirow{30}[6]{*}{CNN1D} & \multirow{10}[2]{*}{RML2016.10a} & RFP   & Channel & 60.02 & 58.93 & -1.09 ↓ & -7.14M & -72.47K & 74.95\% & 71.81\% \\
          &       & FPGM  & Channel & 59.93 & 56.85 & -3.08 ↓ &    -   & -62.67K &     -  & 62.11\% \\
          &       & L1-norm & Channel & 59.93 & 56.95 & -2.98 ↓& -7.14M & -72.47K & 74.95\% & 71.81\% \\
          &       & SFP   & Channel & 59.93 & 57.71 & -2.22 ↓&   -    & -54.13K &    -   & 53.64\% \\
          &       & BNP   & Channel &    -   &    -   &    -   &   -    &   -    &    -   & - \\
          &       & Random & Layer & 59.45 & 57.67 & -1.78 ↓& -0.79M & -53.25K & 8.30\% & 52.80\% \\
          &       & LCP   & Layer & 59.45 & 58.85 & -0.60 ↓ & -0.79M & -53.25K & 8.30\% & 52.80\% \\
          &       & SR-init & Layer & 59.45 & 58.00    & -1.45 ↓& -0.79M & -53.25K & 8.30\% & 52.80\% \\
          &       & PSR   & Layer & 59.45 & 59.18 & -0.27 ↓& -0.79M & -53.25K & 8.30\% & 52.80\% \\
          &       & \cellcolor[rgb]{ .839,  .863,  .894}FCOS  & \cellcolor[rgb]{ .839,  .863,  .894}Channel+Layer & \cellcolor[rgb]{ .839,  .863,  .894}59.45 & \cellcolor[rgb]{ .839,  .863,  .894}59.08 & \cellcolor[rgb]{ .839,  .863,  .894}-0.37 ↓& \cellcolor[rgb]{ .839,  .863,  .894}-7.17M & \cellcolor[rgb]{ .839,  .863,  .894}-66.83K & \cellcolor[rgb]{ .839,  .863,  .894}75.24\% & \cellcolor[rgb]{ .839,  .863,  .894}66.23\% \\
\cmidrule{2-11}          & \multirow{10}[2]{*}{Sig2019-12} & RFP   & Channel & 64.85 & 62.35 & -2.50 ↓ & -28.58M & -72.47K & 74.95\% & 67.47\% \\
          &       & FPGM  & Channel & 64.86 & 60.92 & -3.94 ↓ &    -   & -71.22K &   -    & 56.70\% \\
          &       & L1-norm & Channel & 64.86 & 62.27 & -2.59 ↓& -28.58M & -72.47K & 74.95\% & 67.47\% \\
          &       & SFP   & Channel & 64.86 & 60.96 & -3.90 ↓  &   -    & -71.22K &     -  & 56.70\% \\
          &       & BNP   & Channel &   -    &   -    &    -   &   -    &   -    &   -    &  -\\
          &       & Random & Layer & 64.51 & 59.88 & -4.63 ↓& -3.19M & -53.25K & 8.40\% & 42.40\% \\
          &       & LCP   & Layer & 64.51 & 59.81 & -4.70 ↓  & -3.19M & -53.25K & 8.40\% & 42.40\% \\
          &       & SR-init & Layer & 64.51 & 59.91 & -4.60 ↓  & -3.19M & -53.25K & 8.40\% & 42.40\% \\
          &       & PSR   & Layer & 64.51 & 60.34 & -4.17 ↓& -3.19M & -53.25K & 8.40\% & 42.40\% \\
          &       & \cellcolor[rgb]{ .839,  .863,  .894}FCOS  & \cellcolor[rgb]{ .839,  .863,  .894}Channel+Layer & \cellcolor[rgb]{ .839,  .863,  .894}64.51 & \cellcolor[rgb]{ .839,  .863,  .894}63.30 & \cellcolor[rgb]{ .839,  .863,  .894}-1.21 ↓& \cellcolor[rgb]{ .839,  .863,  .894}-28.62M & \cellcolor[rgb]{ .839,  .863,  .894}-71.69K & \cellcolor[rgb]{ .839,  .863,  .894}75.05\% & \cellcolor[rgb]{ .839,  .863,  .894}57.08\% \\
\cmidrule{2-11}          & \multirow{10}[2]{*}{RML2018.01a} & RFP   & Channel & 87.78 & 79.70  & -8.08 ↓& -57.16M & -101.1K & 74.95\% & 63.23\% \\
          &       & FPGM  & Channel & 87.28 & 82.89 & -4.39 ↓ &    -   & -71.22K &   -    & 44.53\% \\
          &       & L1-norm & Channel & 87.28 & 82.50  & -4.78 ↓& -57.16M & -101.1K & 74.95\% & 63.23\% \\
          &       & SFP   & Channel & 87.28 & 80.78 & -6.50 ↓ &   -    & -62.67K &     -  & 39.19\% \\
          &       & BNP   & Channel &    -   &    -   &    -   &    -   &   -    &     -  & - \\
          &       & Random & Layer & 84.16 & 75.89 & -8.27 ↓& -6.38M & -53.25K & 8.40\% & 33.30\% \\
          &       & LCP   & Layer & 84.16 & 80.93 & -3.23 ↓& -6.38M & -53.25K & 8.40\% & 33.30\% \\
          &       & SR-init & Layer & 84.16 & 81.90  & -2.26 ↓ & -6.38M & -53.25K & 8.40\% & 33.30\% \\
          &       & PSR   & Layer & 84.16 & 82.69 & -1.47 ↓& -6.38M & -53.25K & 8.40\% & 33.30\% \\
          &       & \cellcolor[rgb]{ .839,  .863,  .894}FCOS  & \cellcolor[rgb]{ .839,  .863,  .894}Channel+Layer & \cellcolor[rgb]{ .839,  .863,  .894}84.16 & \cellcolor[rgb]{ .839,  .863,  .894}84.74 & \cellcolor[rgb]{ .839,  .863,  .894}+0.58 ↑ & \cellcolor[rgb]{ .839,  .863,  .894}-57.23M & \cellcolor[rgb]{ .839,  .863,  .894}-71.7K & \cellcolor[rgb]{ .839,  .863,  .894}75.05\% & \cellcolor[rgb]{ .839,  .863,  .894}44.83\% \\
    \bottomrule
    \end{tabular}}
  \label{tab:sota}%
\end{table*}%

\subsection{Models and Datasets Used}
\textbf{Datasets.} We conduct experiments on three signal modulation classification datasets RML2016.10a~\cite{o2016radio}, Sig2019-12~\cite{chen2021signet} and RML2018.01a~\cite{o2018over}. Herein, we elaborate on the description of these datasets.

RML2016.10a is a signal dataset generated by GNU Radio~\cite{blossom2004gnu}. The dataset contains $11$ different signal modulation types and a total of $220,000$ signal samples. $1,000$ samples per modulation type per SNR, signal length is uniformly $128$ sampling points. The SNR range covers $-20$dB to $+18$dB, with a total of $20$ SNR levels in $2$dB intervals. The dataset is divided into training set, validation set, and testing set in a ratio of $6:2:2$ to facilitate the training and performance evaluation of deep learning models.

Sig2019-12 is generated by Chen et al.~\cite{chen2021signet} through simulation, it contains $12$ types of modulation signals. The signal length is uniformly $512$ sampling points. The SNR spans from $-20$dB to $+30$dB in $2$dB intervals, and the total number of samples is $468,000$. $1, 500$ samples per modulation type of each SNR. The dataset is divided into training set, validation set and testing set in a ratio of $6:2:2$.

RML2018.01a simulates a wireless channel and is collected from a laboratory environment. It contains $24$ modulation types. The signal length is extended to $1,024$ sampling points, and the SNR range is also $-20$dB to $+30$dB in $2$dB intervals. Since the total amount of data exceeds $2.55$ million, only samples with a SNR of more than $10$dB are selected in the experiment, and divided into training set and testing set in a ratio of $8:2$.

\textbf{Models.} We use typical convolutional neural networks designed for the AMR task (ResNet56~\cite{he2016deep}, CNN1D~\cite{o2018over}) and SigNet50~\cite{chen2021signet} as base models for pruning.

\subsection{Implementation details}
For a fair comparison, we set the same hyperparameter settings to fine-tune pruned models. As for FCOS, in the fine-tuning stage of channel pruning, we use a learning rate of $0.001$ and a batch size of $128$ for $20$ epochs. Then in the linear probe stage, we also use a learning rate of $0.001$ and a batch size of $128$, but only train for $5$ epochs. Finally, in the fine-tuning stage of layer pruning, we still keep the learning rate of $0.001$ and the batch size of $128$, but increase the training rounds to $80$ epochs to fully recover the performance of the pruned model. As for baselines, we set the initial learning rate, batch size and epoch to 0.001, 128 and 80, respectively.

\subsection{Baselines}
To verify the effectiveness of our method, we compare it with channel pruning methods RFP~\cite{shao2021filter}, FPGM~\cite{he2019filter}, L1-norm~\cite{li2016pruning}, SFP~\cite{he2018soft} and BNP~\cite{chen2023channel}, as well as layer pruning methods random, LCP~\cite{chen2018shallowing,wang2019dbp}, PSR~\cite{lu2024generic} and SR-init~\cite{tang2023sr}. A detailed description of these methods is given in \cref{tab:baseline intro}. It is worth noting that FPGM and SFP are soft pruning, which only zeros some weights during pruning, but does not actually change the model structure (channels still remain in the model). Therefore, they will not bring substantial runtime acceleration. Since both SFP and FPGM use a soft pruning strategy and do not actually remove channels, their FLOPs will not be reduced. Hence, in the following comparison, we will no longer count the FLOPs changes of these two methods, but only focus on their parameter compression effect and accuracy performance. Besides, since BNP specifically operates on the scaling factor of the  batch normalization, and there is no BatchNorm layer in CNN1D, we do not consider using BNP for pruning on the CNN1D model.

\vspace{-3mm}
\subsection{Results and Analysis}
\label{sec:Results and Analysis}
\begin{figure*}[t]
	\centering
    \includegraphics[width=0.99\textwidth]{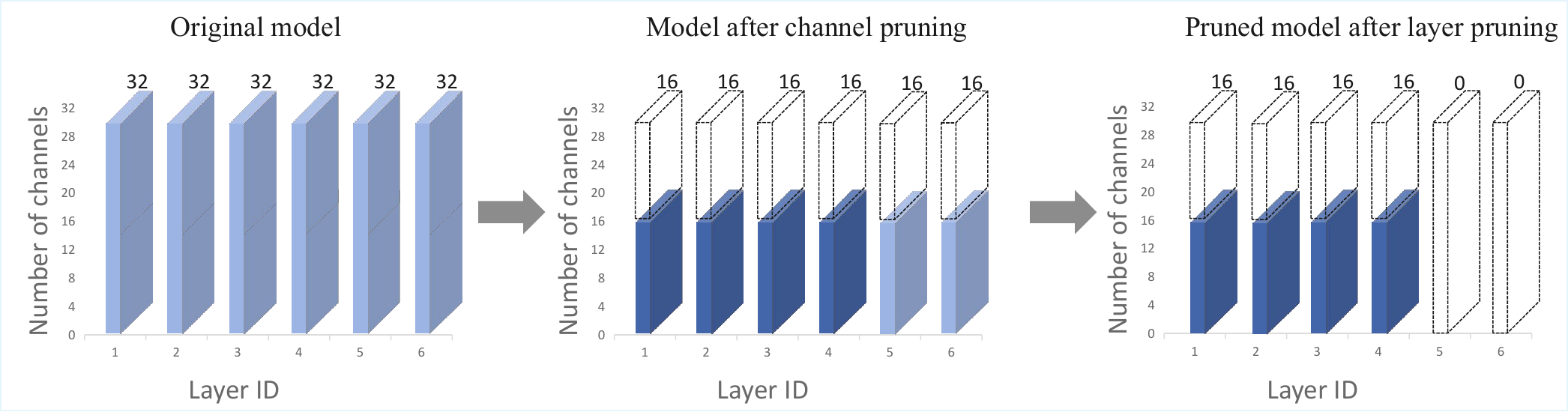} 
 \caption{Visualization of the changes in layers and channels during pruning of CNN1D on RML2016.10a.}
 \label{fig:vis_layer}
\end{figure*}

\textbf{Comparison to channel pruning methods.} In this paragraph, we compare FCOS with channel pruning methods on both three AMR datasets. As shown in \cref{tab:sota}, when the pruning rate is extremely high, FCOS can always maintain or even slightly improve the performance of the original model, while other methods generally show a significant drop in accuracy. For example, the accuracy of the original ResNet56 is 62.06\% on RML2016.10a. After FCOS pruning, the FLOPs are reduced by 88.39\%, the number of parameters is reduced by 88.34\%, and the accuracy is improved to 62.12\% (+0.06\%). In contrast, typical channel pruning methods (such as RFP, FPGM, L1‑norm, SFP and BNP) have a similar reduction in FLOPs and Params, but the accuracy drops by 1.24\% to 5.98\%. For Sig2019-12, the original ResNet56 has an accuracy of 67.91\%. After FCOS pruning, FLOPs are reduced by 95.51\%, the number of parameters is reduced by 95.31\%, and the accuracy only drops slightly to 67.45\% (-0.46\%). In contrast, the typical channel pruning methods achieve a similar FLOPs reduction and less parameter reduction, but the accuracy drops by 3.03\% (L1-norm) to 49.00\% (SFP). For RML2018.01a, the original ResNet56 has an accuracy of 89.70\%. After FCOS pruning, FLOPs are reduced by 90.09\%, the number of parameters is reduced by 90.03\%, and the accuracy is improved to 90.61\% (+0.91\%). In contrast, traditional channel pruning methods achieve a higher FLOPs reduction and similar parameter reduction, but the accuracy drops by 9.21\% to 59.20\%. The same situation is also observed when pruning with CNN1D on these datasets, which demonstrates the superiority of FCOS over existing channel pruning methods. Besides, we also provide the changes in layers and channels during pruning of CNN1D on RML2016.10a in \cref{fig:vis_layer}.

\begin{table*}[t]
  \centering
  \caption{PRUNING RESULTS OF SigNet50 ON RML2016.10a and Sig2019-12. PR DENOTES THE PRUNING RATE.}
    \begin{tabular}{cccccccccc}
    \toprule
    Dataset & Method & Pruning Type &  Original Acc(\%) &  Acc(\%) &  $\triangle$Acc(\%) &  $\triangle$FLOPs &  $\triangle$Params &  Params PR &  FLOPs PR \\
    \midrule
    \multirow{5}[2]{*}{RML2016.10a} & Random & Layer &  63.41 &  62.85 &  -0.56 ↓ &  -0.80G &  -15.21M &  64.60\% &  60.30\% \\
          & LCP   & Layer &  63.41  &  62.49  &  -0.92 ↓  &  -0.80G  &  -14.38M  &  61.10\%  &  60.30\%  \\
          & SR-init & Layer &  63.41  &  62.54  &  -0.87 ↓  &  -0.80G  &  -14.38M  &  61.10\%  &  60.30\%  \\
          & PSR   & Layer &  63.41  &  62.94  &  -0.47 ↓  &  -0.80G  &  -15.42M  &  65.50\%  &  60.30\%  \\
          &  \cellcolor[rgb]{ .839,  .863,  .894}FCOS  &  \cellcolor[rgb]{ .839,  .863,  .894}Channel+Layer &  \cellcolor[rgb]{ .839,  .863,  .894}63.41 & \cellcolor[rgb]{ .839,  .863,  .894}62.33 &  \cellcolor[rgb]{ .839,  .863,  .894}-1.08 ↓ &  \cellcolor[rgb]{ .839,  .863,  .894}-1.29G &  \cellcolor[rgb]{ .839,  .863,  .894}-22.66M &  \cellcolor[rgb]{ .839,  .863,  .894}96.30\% &  \cellcolor[rgb]{ .839,  .863,  .894}95.93\% \\
    \midrule
    \multirow{5}[2]{*}{Sig2019-12} & Random & Layer &  71.47  &  70.11  &  -1.36 ↓  &  -13.75G  &  -14.23M  &  60.50\%  &  64.70\%  \\
          & LCP   & Layer &  71.47  &  70.21  &  -1.26 ↓  &  -13.75G  &  -14.31M  &  60.80\%  &  64.70\%  \\
          & SR-init & Layer &  71.47  &  70.36  &  -1.11 ↓  &  -13.75G  &  -14.38M  &  61.10\%  &  64.70\%  \\
          & PSR   & Layer &  71.47  &  70.36  &  -1.11 ↓  &  -13.75G  &  -15.49M  &  65.80\%  &  64.70\%  \\
          & \cellcolor[rgb]{ .839,  .863,  .894}FCOS  & \cellcolor[rgb]{ .839,  .863,  .894}Channel+Layer &   \cellcolor[rgb]{ .839,  .863,  .894}71.47    &   \cellcolor[rgb]{ .839,  .863,  .894}70.67    &    \cellcolor[rgb]{ .839,  .863,  .894}-0.80 ↓  &  \cellcolor[rgb]{ .839,  .863,  .894}-20.51G     &   \cellcolor[rgb]{ .839,  .863,  .894}-22.66M    &    \cellcolor[rgb]{ .839,  .863,  .894}96.30\%   &  \cellcolor[rgb]{ .839,  .863,  .894}95.93\% \\
    \bottomrule
    \end{tabular}%
  \label{tab:SigNet50}%
\end{table*}%

\textbf{Comparison to layer pruning methods.} In the previous paragraph, we demonstrate the superiority of FCOS compared to existing channel pruning methods. Here, we make a comprehensive comparison with layer pruning methods such as Random, LCP, SR-init and PSR. As shown in \cref{tab:sota}, all layer pruning baselines achieve comparable compression on RML2016.10a, but their accuracies uniformly drop by 0.96\% to 1.98\%. In contrast, FCOS attains an even higher compression rate (88.39\% FLOPs, 88.34\% parameters) while slightly improving accuracy to 62.12\% (+0.06\%). As for Sig2019‑12, all layer pruning baselines deliver nearly identical compression (90.70\% FLOPs reduction and 91.20\% parameter reduction), yet their accuracies fall by 2.98\% to 6.14\%. In contrast, FCOS achieves even greater compression (95.51\% FLOPs, 95.31\% parameters) while keeping accuracy almost unchanged at 67.45\% (–0.46\%). In RML2018.01a, FCOS and other layer pruning methods have almost the same pruning rate, but the accuracy of other methods has dropped significantly by 5.63\% to 7.00\%. In contrast, the accuracy of FCOS has increased by +0.91\% (90.61\%). We observe the same trend when pruning CNN1D in these datasets, further emphasizing the advantage of FCOS over traditional layer pruning methods.

\textbf{Results on more complex models.} In the previous paragraph, we have demonstrated the superiority of FCOS compared to existing channel and layer pruning methods on ResNet56 and CNN1D. In order to prove that FCOS can be used in a variety of model structures, we further perform FCOS pruning on RML2016.10a and Sig2019-12 datasets using SigNet50~\cite{chen2021signet}. As shown in \cref{tab:SigNet50}, FCOS consistently outperforms pure layer pruning on SigNet50 across both datasets: For RML2016.10a, all layer pruning baselines only achieve 60.30\% FLOPs reduction and less than 70\% parameter reduction, with accuracy drops ranging from -0.47\% to -0.92\%. In contrast, FCOS compresses the model more aggressively, reducing FLOPs by 95.93\% and parameters by 96.30\%, while accuracy remains within -1.08\% of the original value. As for Sig2019‑12, layer pruning baselines removes 64.70\% of FLOPs and about 65.00\% of parameters, but this comes at the cost of a 1.11\%-1.36\% drop in accuracy. FCOS reduces FLOPs by 95.93\% and parameters by 96.30\%, while accuracy remains within -0.80\% of the original value. These results achieved on more complex architectures demonstrate that FCOS can achieve extreme pruning and strong accuracy preservation across a variety of model structures.

\begin{figure}[t]
	\centering
    \includegraphics[width=0.49\textwidth]{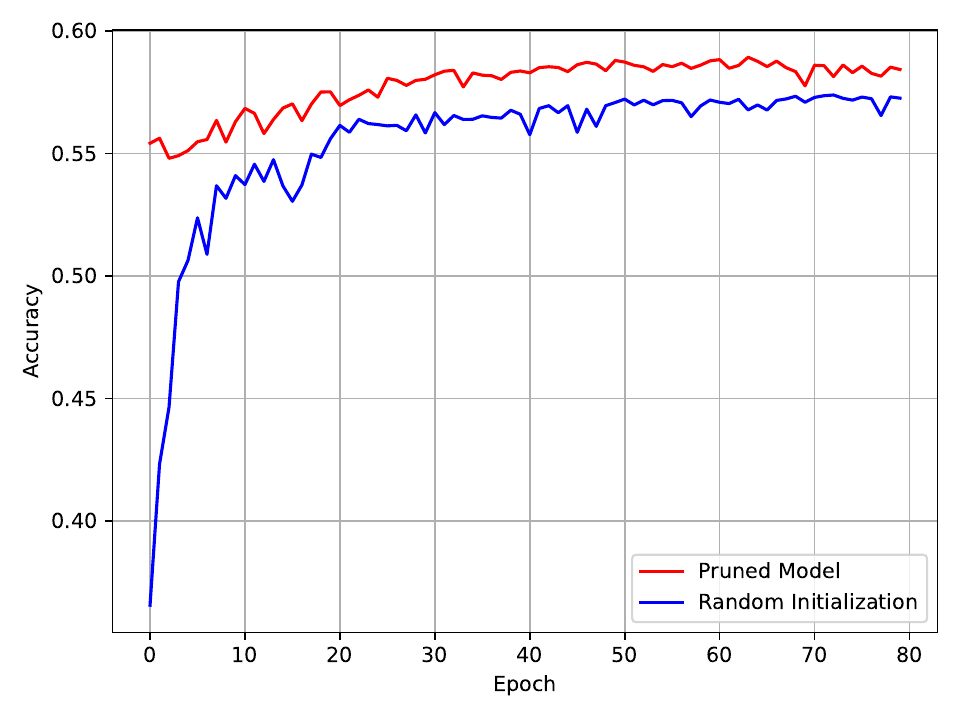} 
 \caption{Comparison of the accuracy curve of the model obtained by FCOS and a model of the same size trained from scratch. Experiments are conducted on RML2016.10a using CNN1D.}
 \label{fig:acc_curve}
\end{figure}



\subsection{Additional Experiments and Analyses}
\label{sec:Additional Experiments and Analyses}
\textbf{Fine-tuning vs. training from scratch.} In this paper, we fine-tune the pruned model by retaining some parameters from the original pre-trained model, instead of randomly initializing all parameters (i.e., training from scratch). To evaluate the effectiveness of such fine-tuning, we compare the accuracy curve of the model obtained by FCOS and a model of the same size trained from scratch. Experiments are conducted on RML2016.10a using CNN1D. As shown in \cref{fig:acc_curve}, loading some weights of the original model for fine-tuning can not only significantly accelerate model convergence, but also improve the final accuracy.

\subsection{Ablation Study}
\label{sec: Ablation Study}
In this subsection, we conduct detailed ablation experiments on the similarity measure required for weight fusion, different weighting methods and the order of operations for weight fusion.

\textbf{The similarity measure required for weight fusion.} As mentioned in \cref{sec:Channel Pruning via Similar Channel Fusion}, we uniformly use cosine similarity 
to calculate the similarity. Here, we want to prove through experiments that any other similarity measure can be used for calculation. Therefore, we choose Euclidean similarity as an alternative metric to conduct experiments, which can be simply defined as:
\begin{equation}
    s_{Euc}=\frac{1}{1+d_{Euc}(x,y)}, \quad \text{where} \quad d_{Euc}(x,y) = ||x-y||_{2}.
    \label{eq:euc sim}
\end{equation} 
Specifically, we conduct experiments on RML2016.10a using CNN1D. As shown in \cref{tab:diff sim}, the cosine-based fusion achieves an accuracy of 59.08\%, while the Euclidean-based fusion slightly outperforms it with an accuracy of 59.39\%, an improvement of 0.31\%. This demonstrates that FCOS is robust to the choice of similarity metric and that its performance does not hinge on a particular similarity measure.

\textbf{The impact of different weighting methods on the performance of pruned models.} In \cref{sec:Channel Pruning via Similar Channel Fusion}, we use the simply averaging (\cref{eq:fusion}) to perform weight fusion. To study the impact of weight fusion methods on pruning performance, we compared different channel weighting schemes. Specifically, we assign merging weights based on each channel’s $L_1$-norm. Concretely, let a given cluster $\mathcal{C}$ (in the $i$-th layer) contain $n$ channels with weight tensors $\{w_{i,1}, w_{i,2},\cdots,w_{i,n}\}$ We compute each channel’s importance score by its $L_1$-norm:
\begin{equation}
s_j=\left\|w_{i,j}\right\|_1, \quad j=1, \ldots, n.
\end{equation}
Then we normalize these scores to obtain fusion weights:
\begin{equation}
\alpha_j=\frac{s_j}{\sum_{k=1}^n s_k}, \quad \sum_{i=1}^n \alpha_j=1.
\end{equation}
The merged cluster weight $w_i^*$ is then formed by the weighted sum:
\begin{equation}
\widetilde{w}=\sum_{j=1}^n \alpha_j w_{i,j}.
\label{eq:l1norm}
\end{equation}
Then we use \cref{eq:fusion} and \cref{eq:l1norm} to conduct weight fusion. As shown in \cref{tab:fusion method}, we find that the performance difference of the models after pruning with two different fusion methods is very small, which demonstrates that FCOS is robust to the choice of fusion method.

\textbf{The order of weight fusion operations.} As mentioned in \cref{sec:Channel Pruning via Similar Channel Fusion}, we first perform channel fusion on the output channels and then repeat it on the input channels. Here, we want to prove through experiments that the fusion order has no effect on the pruning effect. We swap the fusion order, that is, first perform input channel fusion, then output channel fusion, and other operations remain unchanged. Specifically, we conduct experiments on RML2016.10a using CNN1D. As shown in \cref{tab:fusion order}, the order of the weight fusion operation — whether it is applied to the input channels first or the output channels first — does have a minimal impact on the performance of the pruning model.

\begin{table}[t]
  \centering
  \caption{The effect of the order of weight fusion operations on FCOS.}
    \begin{tabular}{ccccc}
    \toprule
    Fusion Type & Original Acc(\%) & Acc(\%) & Params PR & FLOPs PR \\
    \midrule
    Output - Input & 59.45 & 59.08 & 66.23\% & 75.24\% \\
     Input - Output & 59.45 & 59.03 & 66.23\% & 75.24\% \\
    \bottomrule
    \end{tabular}%
  \label{tab:fusion order}%
\end{table}%

\begin{table}[t]
  \centering
  \caption{The impact of different weighting methods on the performance of pruned models.}
    \begin{tabular}{ccccc}
    \toprule
    Fusion Method & Original Acc(\%) & Acc(\%) & Params PR & FLOPs PR \\
    \midrule
       \cref{eq:fusion}   & 59.45 & 59.08 & 66.23\% & 75.24\% \\
       \cref{eq:l1norm}   & 59.45 & 58.83 & 66.23\% & 75.24\% \\
    \bottomrule
    \end{tabular}%
  \label{tab:fusion method}%
\end{table}%

\begin{table}[t]
  \centering
  \caption{The impact of different similarity measures required for weight fusion on the performance of the pruned model. The definition of euclidean similarity is shown in \cref{eq:euc sim}.}
  \resizebox{0.49\textwidth}{!}{
    \begin{tabular}{ccccc}
    \toprule
    Similarity Index & Original Acc(\%) & Acc(\%) & Params PR & FLOPs PR \\
    \midrule
     Cosine Similarity & 59.45 & 59.08 & 66.23\% & 75.24\% \\
    Euclidean Similarity & 59.45 & 59.39 & 66.23\% & 75.24\% \\
    \bottomrule
    \end{tabular}}
  \label{tab:diff sim}%
\end{table}%

\section{Conclusion}
\label{sec:Conclusion}
In this paper, we propose FCOS, a fine-to-coarse two-stage pruning framework that combines channel-level pruning with layer-level collapse diagnosis to achieve extreme compression, high performance and efficient inference. Specifically, FCOS consists of two stages. In the first stage, hierarchical clustering and parameter fusion are applied to channel weights to achieve channel-level pruning. Then, Layer Collapse Diagnosis (LaCD) module utilizes linear probing to diagnose layer collapse and removes these layers. Experiments on multiple AMR datasets demonstrate that FCOS outperforms existing channel and layer pruning methods while achieving extremely high compression rates.

%
\bibliographystyle{IEEEtran}
\bibliography{reference}

\begin{IEEEbiography}[{\includegraphics[width=1in,height=1.25in,clip,keepaspectratio]{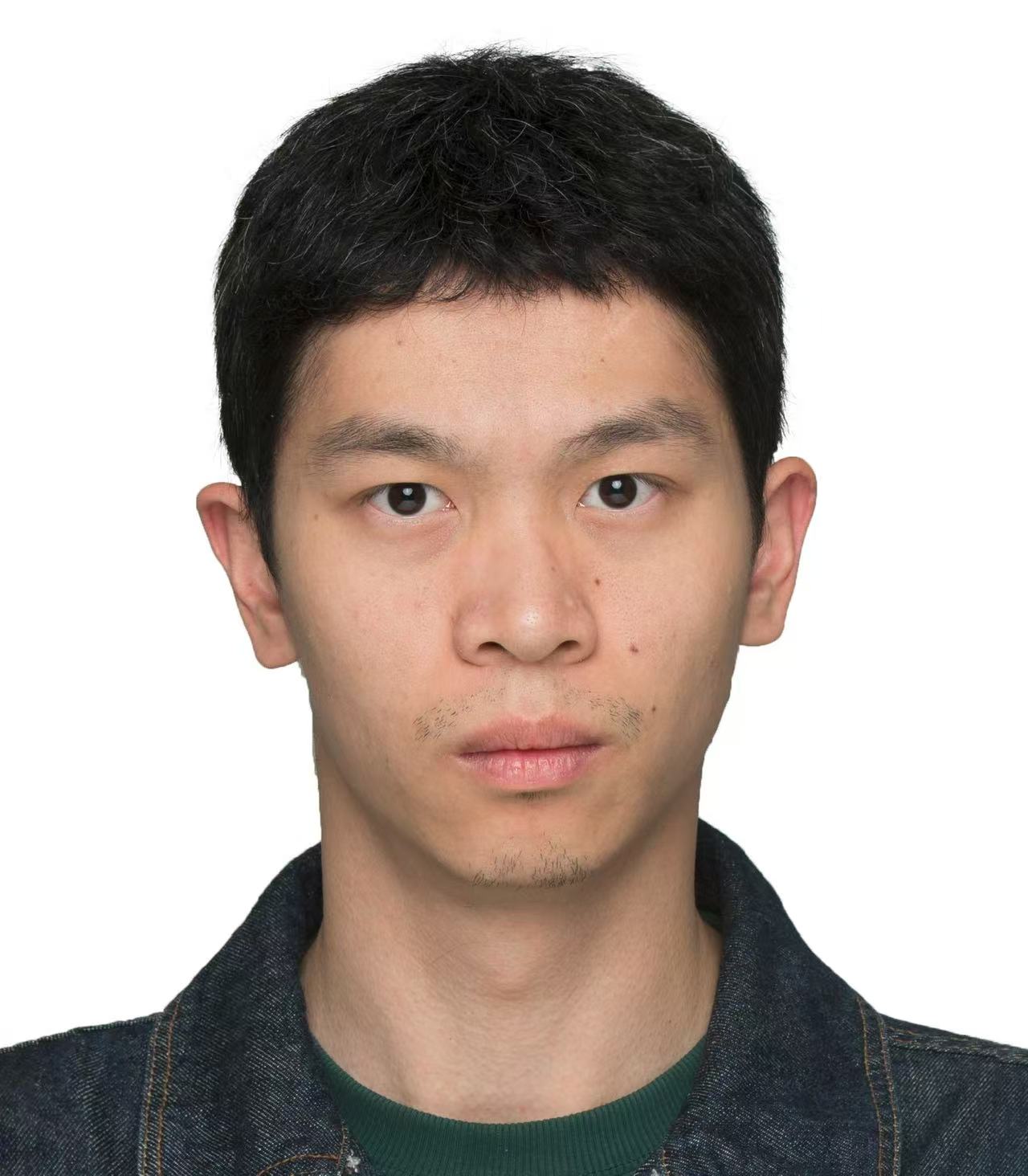}}]{Yao Lu}
received his B.S. degree from Zhejiang University of Technology and is currently pursuing a Ph.D. in control science and engineering at Zhejiang University of Technology. He is a visiting scholar with the Centre for Frontier AI Research, Agency for Science, Technology and Research, Singapore. He has published several academic papers in international conferences and journals, including ECCV, TNNLS, Neurocomputing and TCCN. He serves as a reviewer of ICLR2025, CVPR2025, ICCV25 and TNNLS. His research interests include deep learning and computer vision, with a focus on artificial intelligence and model compression.
\end{IEEEbiography} 
\vspace{-15mm}

\begin{IEEEbiography}[{\includegraphics[width=1in,height=1.25in,clip,keepaspectratio]{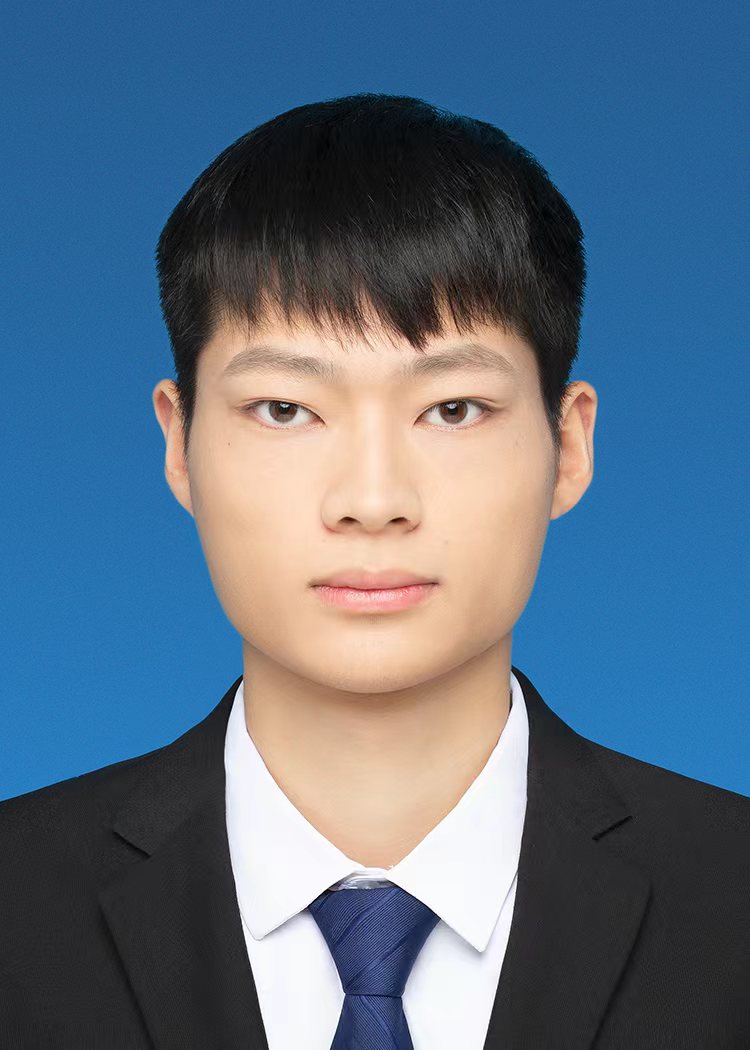}}]{Tengfei Ma}
received his bachelor’s degree from the School of Electrical and Electronic Engineering at Hangzhou City University in 2024. He is currently a master’s student at the School of Information Engineering at Zhejiang University of Technology. His research interests include signal processing and lightweight neural networks.
\end{IEEEbiography} 

\vspace{-15mm}
\begin{IEEEbiography}[{\includegraphics[width=1in,height=1.25in,clip,keepaspectratio]{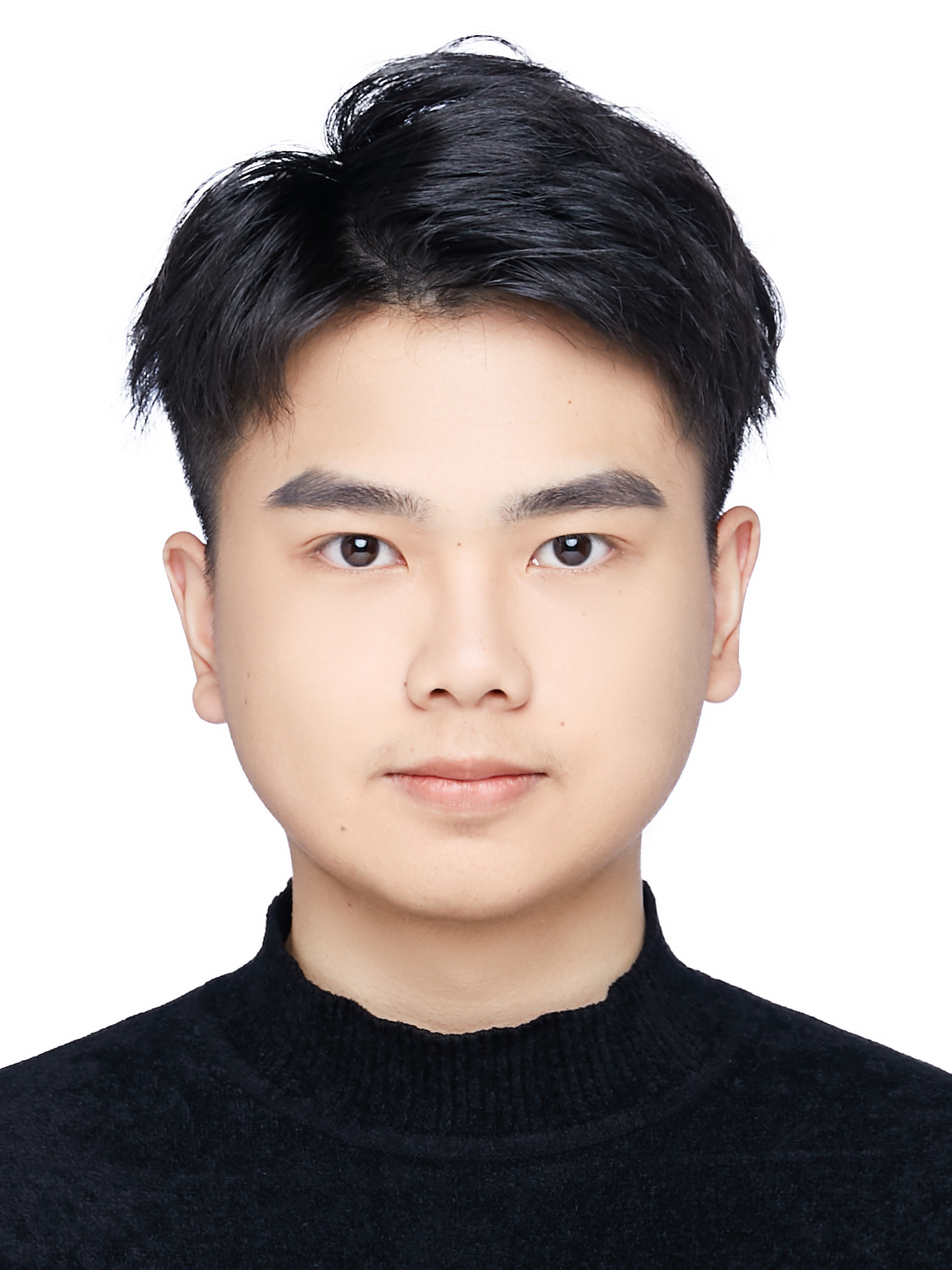}}]{Zeyu Wang} received the BS degree from Shaoxing University, Shaoxing, China, in 2021. He is currently pursuing the Ph.D. degree at the College of Information Engineering, Zhejiang University of Technology, Hangzhou, China. His current research interests include graph data mining, recommender system and bioinformatics.
\end{IEEEbiography}

\vspace{-15mm}

\begin{IEEEbiography}[{\includegraphics[width=1in,height=1.25in,clip,keepaspectratio]{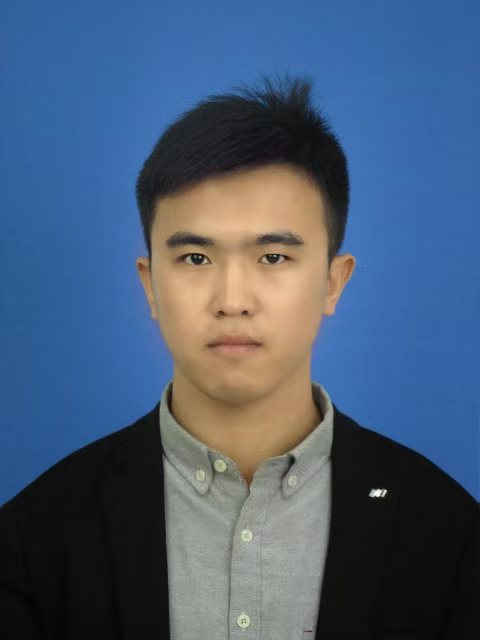}}]{Zhuangzhi Chen}
received the B.S. and Ph.D. degrees in Control Theory and Engineering from Zhejiang University of Technology, Hangzhou, China, in 2017 and 2022, respectively. He was a visiting scholar with the Department of Computer Science, University of California at Davis, CA, in 2019. He completed his postdoctoral research in Computer Science and Technology at Zhejiang University of Technology, in 2024. He is currently working as an associate researcher at the Cyberspace Security Research Institute of Zhejiang University of Technology, and also serving as the deputy director and associate researcher at the Research Center of Electromagnetic Space Security, Binjiang Institute of Artificial Intelligence, ZJUT, Hangzhou 310056, China.

His current research interests include deep learning algorithm, deep learning-based radio signal recognition, and security AI of wireless communication.
\end{IEEEbiography} 
\vspace{-15mm}
\begin{IEEEbiography}[{\includegraphics[width=1in,height=1.25in,clip,keepaspectratio]{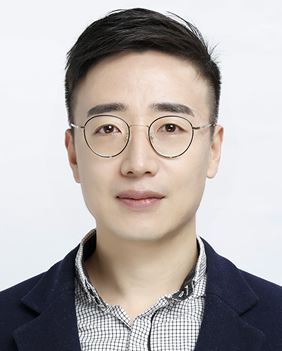}}]{Dongwei Xu}
(Member, IEEE) received the B.E. and Ph.D. degrees from the State Key Laboratory of Rail Traffic Control and Safety, Beijing Jiaotong University, Beijing, China, in 2008 and 2014, respectively. He is currently an Associate Professor with the Institute of Cyberspace Security, Zhejiang University of Technology, Hangzhou, China. His research interests include intelligent transportation Control, management, and traffic safety engineering.
\end{IEEEbiography}

\begin{IEEEbiography}[{\includegraphics[width=1in,height=1.25in,clip,keepaspectratio]{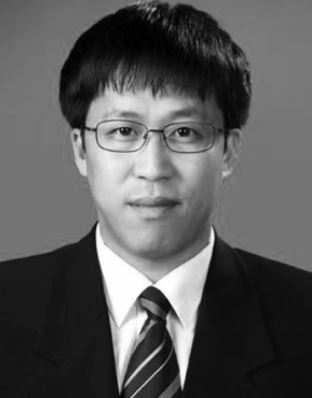}}]{Yun Lin}
(Member, IEEE) received the B.S. degree in electrical engineering from Dalian Maritime University, Dalian, China, in 2003, the M.S. degree in communication and information system from the Harbin Institute of Technology, Harbin, China, in 2005, and the Ph.D. degree in communication and information system from Harbin Engineering University, Harbin, in 2010. From 2014 to 2015, he was a Research Scholar with Wright State University, Dayton, OH, USA. He is currently a Full Professor with the College of Information and Communication Engineering, Harbin Engineering University. He has authored or coauthored more than 200 international peer-reviewed journal/conference papers, such as IEEE Transactions on Industrial Informatics, IEEE Transactions on Communications, IEEE Internet of Things Journal, IEEE Transactions on Vehicular Technology, IEEE Transactions on Cognitive Communications and Networking, TR, INFOCOM, GLOBECOM, ICC, VTC, and ICNC. His current research interests include machine learning and data analytics over wireless networks, signal processing and analysis, cognitive radio and software-defined radio, artificial intelligence, and pattern recognition.
\end{IEEEbiography}

\begin{IEEEbiography}[{\includegraphics[width=1in,height=1.25in,clip,keepaspectratio]{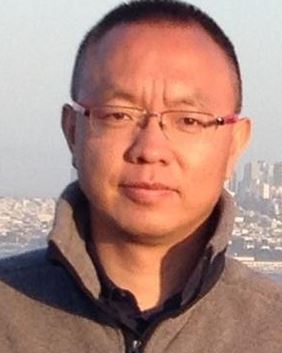}}]{Qi Xuan}
(Senior Member, IEEE) received the B.S. and Ph.D. degrees in control theory and engineering from Zhejiang University, Hangzhou, China, in 2003 and 2008, respectively. He was a Postdoctoral Researcher with the Department of Information Science and Electronic Engineering, Zhejiang University from 2008 to 2010, and a Research Assistant with the Department of Electronic Engineering, City University of Hong Kong, Hong Kong, in 2010 and 2017, respectively. From 2012 to 2014, he was a Postdoctoral Fellow with the Department of Computer Science, University of California at Davis, Davis, CA, USA. He is currently a Professor with the Institute of Cyberspace Security, College of Information Engineering, Zhejiang University of Technology, Hangzhou, and also with the PCL Research Center of Networks and Communications, Peng Cheng Laboratory, Shenzhen, China. He is also with Utron Technology Company Ltd., Xi’an, China, as a Hangzhou Qianjiang Distinguished Expert. His current research interests include network science, graph data mining, cyberspace security, machine learning, and computer vision.
\end{IEEEbiography}

\begin{IEEEbiography}[{\includegraphics[width=1in,height=1.25in,clip,keepaspectratio]{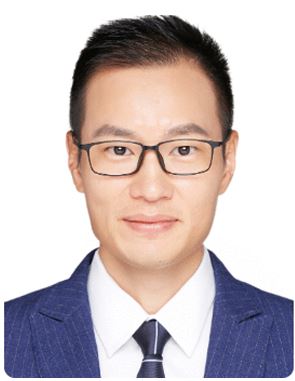}}]{Guan Gui}
    received the Dr. Eng. degree in information and communication engineering from the University of Electronic Science and Technology of China, Chengdu, China, in 2012. From 2009 to 2012, with financial support from the China Scholarship Council and the Global Center of Education, Tohoku University, he joined the Wireless Signal Processing and Network Laboratory (Prof. Adachis laboratory), Department of Communications Engineering, Graduate School of Engineering, Tohoku University, as a Research Assistant and a Post Doctoral Research Fellow, respectively. From 2012 to 2014, he was supported by the Japan Society for the Promotion of Science Fellowship as a Post Doctoral Research Fellow with the Wireless Signal Processing and Network Laboratory. From 2014 to 2015, he was an Assistant Professor with the Department of Electronics and Information System, Akita Prefectural University. Since 2015, he has been a Professor with the Nanjing University of Posts and Telecommunications, Nanjing, China. He is currently involved in the research of big data analysis, multidimensional system control, super-resolution radar imaging, adaptive filter, compressive sensing, sparse dictionary designing, channel estimation, and advanced wireless techniques. He received the IEEE International Conference on Communications Best Paper Award in 2014 and 2017 and the IEEE Vehicular Technology Conference (VTC-spring) Best Student Paper Award in 2014. He was also selected as a Jiangsu Special Appointed Professor, as a Jiangsu High-Level Innovation and Entrepreneurial Talent, and for 1311 Talent Plan in 2016. He has been an Associate Editor of the Wiley Journal Security and Communication Networks since 2012.
\end{IEEEbiography}





\vfill

\end{document}